%% file: ieee_vr_2020.tex
\documentclass[preprint]{vgtc}               

\usepackage[colorlinks=true,bookmarks=false]{hyperref}
\ifpdf
  \pdfoutput=1\relax                   
  \usepackage{graphicx}                
  \DeclareGraphicsExtensions{.pdf,.png,.jpg,.jpeg} 
\else
  \usepackage{graphicx}                
  \DeclareGraphicsExtensions{.eps}     
\fi%

\graphicspath{{figures/}{pictures/}{images/}{./}} 

\usepackage{microtype}                 
\PassOptionsToPackage{warn}{textcomp}  
\usepackage{textcomp}                  
\usepackage{mathptmx}                  
\usepackage{times}                     
\usepackage{cite}                      
\usepackage{tabu}                      
\usepackage{booktabs}                  
\usepackage{amsmath}
\usepackage{amssymb}
\usepackage{newlfont}
\usepackage{MnSymbol}
\usepackage{balance}
\usepackage{graphicx}
\usepackage{multirow}
\usepackage{wasysym}
\usepackage{comment}
\usepackage[font=small,labelfont=bf,tableposition=top]{caption}
\DeclareMathAlphabet{\pazocal}{OMS}{zplm}{m}{n}
\usepackage{pifont}
\newcommand{\cmark}{\checked}%
\newcommand{\xmark}{}

\vgtcinsertpkg

\preprinttext{To appear in an IEEE VGTC sponsored conference.}

\title{Deep Soft Procrustes for Markerless Volumetric Sensor Alignment}

\author{Vladimiros Sterzentsenko\thanks{e-mail: \href{mailto:vladster@iti.gr}{vladster@iti.gr} -- Soft procrustes implementation, experiments design, implementation and conducting} %
\and Alexandros Doumanoglou\thanks{e-mail: \href{mailto:aldoum@iti.gr}{aldoum@iti.gr} -- Rendering data generation implementation, experiments design, metric definition and implementation} %
\and Spyridon Thermos\thanks{e-mail: \href{mailto:spthermo@iti.gr}{spthermo@iti.gr} -- Network and soft attention design and implementation}
\and Nikolaos Zioulis\thanks{e-mail: \href{mailto:nzioulis@iti.gr}{nzioulis@iti.gr} -- Concept formulation, original rendering implementation and multi-view dense refinement optimization implementation}%
\and Dimitrios Zarpalas\thanks{e-mail: \href{mailto:zarpalas@iti.gr}{zarpalas@iti.gr} -- Shared senior authorship}%
\and Petros Daras\thanks{e-mail: \href{mailto:daras@iti.gr}{daras@iti.gr} -- Shared senior authorship}%
}
\affiliation{\scriptsize Visual Computing Lab, Information Technologies Institute, Centre for Research and Technology Hellas}

\abstract{
With the advent of consumer grade depth sensors, low-cost volumetric capture systems are easier to deploy. 
Their wider adoption though depends on their usability and by extension on the practicality of spatially aligning multiple sensors.
Most existing alignment approaches employ visual patterns, \textit{e.g.}~checkerboards, or markers and require high user involvement and technical knowledge. 
More user-friendly and easier-to-use approaches rely on markerless methods that exploit geometric patterns of a physical structure. 
However, current SoA approaches are bounded by restrictions in the placement and the number of sensors.
In this work, we improve markerless data-driven correspondence estimation to achieve more robust and flexible multi-sensor spatial alignment. 
In particular, we incorporate geometric constraints in an end-to-end manner into a typical segmentation based model and bridge the intermediate dense classification task with the targeted pose estimation one. 
This is accomplished by a soft, differentiable procrustes analysis that regularizes the segmentation and achieves higher extrinsic calibration performance in expanded sensor placement configurations, while being unrestricted by the number of sensors of the volumetric capture system. Our model is experimentally shown to achieve similar results with marker-based methods and outperform the markerless ones, while also being robust to the pose variations of the calibration structure.
Code and pretrained models are available at \href{https://vcl3d.github.io/StructureNet/}{https://vcl3d.github.io/StructureNet/}.
} 

\hypersetup{
    colorlinks = true
}
\usepackage{stackengine} 
\usepackage[dvipsnames]{xcolor}

\CCScatlist{
 \CCScatTwelve{\textbf{Computing methodologies}}{\textbf{Artificial intelligence}}{\textbf{Computer vision}}{Image segmentation};
 \CCScatTwelve{\textbf{Computing methodologies}}{\textbf{Artificial intelligence}}{\textbf{Computer vision}}{Camera calibration};
 \CCScatTwelve{\textbf{Computing methodologies}}{\textbf{Artificial intelligence}}{\textbf{Computer vision}}{3D imaging};
}

\begin{document}
\maketitle



\section{Introduction} 
\label{sec:intro}
\input{Introduction.tex}

\section{Related Work}
\label{sec:related}
\input{RelatedWork.tex}

\section{Approach}
\label{sec:approach}
\input{Approach.tex}
\subsection{Semantic Correspondences}
\label{sec:semantic_corrs}
\input{SemanticCorrespondences.tex}

\subsection{Soft Procrustes}
\label{sec:soft_procrustes}
\input{SoftProcrustes.tex}




\section{Experimental Results}
\label{sec:results}
\input{Results.tex}

\vspace{-3pt}

\section{Conclusion}
\label{sec:conclusion}
\input{Conclusion.tex}

\vspace{-2pt}
\acknowledgments{
We acknowledge HW support by NVidia and financial support by the H2020 EC project Hyper360 (GA 761934).}

\balance
\bibliographystyle{abbrv-doi-hyperref}

\bibliography{./bibs/methods,./bibs/pose,./bibs/augmentations,./bibs/features,./bibs/systems,./bibs/sensors,./bibs/applications,./bibs/cnn}
\end{document}

%% file: Introduction.tex
Cameras, as well as range imaging sensors, enable the digitization of real world scenes.
Using multiple spatially aligned sensors is a widely applied and viable approach to volumetrically (\textit{i.e.}~full 3D) capture real scenes in motion. 
Research and technology progress have recently converged to a point where it is possible to comfortably deploy multi-sensor setups for volumetric capturing.
Recent integrated RGB-D sensors \cite{6190806}, as well as the optimization of the stereo algorithms \cite{DBLP:journals/corr/KeselmanWGB17} in combination with high-end GPU processing, have enabled higher quality 3D capturing with lower cost systems.
Be it either of average cost as the Holoportation system introduced in~\cite{Holoportation} that used 16 near-infrared stereo cameras and 8 color texturing units, or less costly solutions relying on cheaper RGB-D sensors \cite{sterzentsenko2018low}, all approaches rely on a precise volumetric alignment\footnote{We use the terms volumetric and spatial alignment interchangeably in this document and they both refer to external (\textit{i.e.}~extrinsics) calibration (\textit{i.e.}~registration) of the system.} of the corresponding sensors.
This drives the subsequent processing that leverages different methods for digitizing human performances either online with non-rigid registration \cite{Fusion4d} or near real-time with template fitting algorithms \cite{alexiadis2018fast}.

\begin{figure}[!htp]
\begin{center}
   \includegraphics[width=0.75\columnwidth]{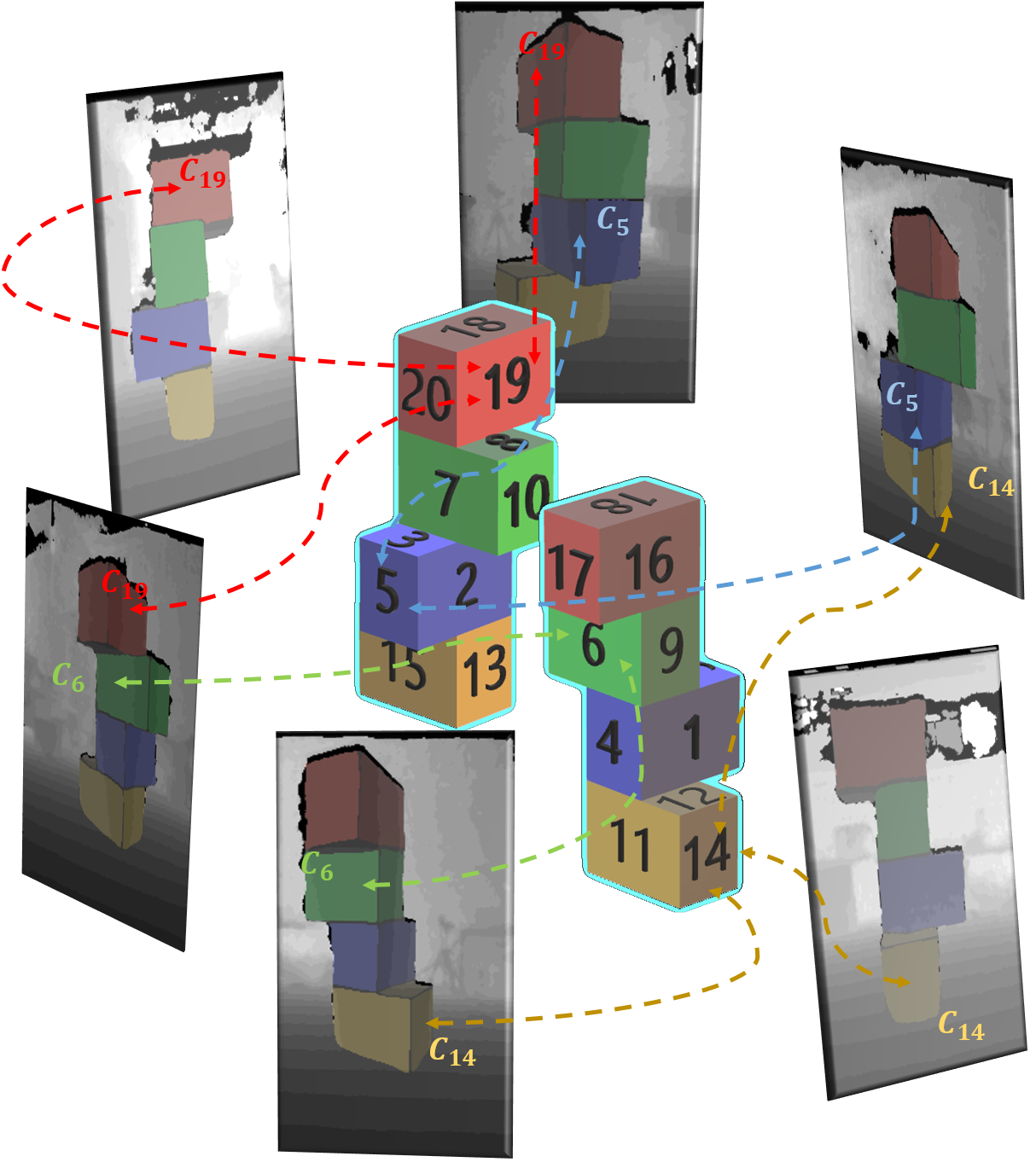}
   \includegraphics[width=0.95\columnwidth]{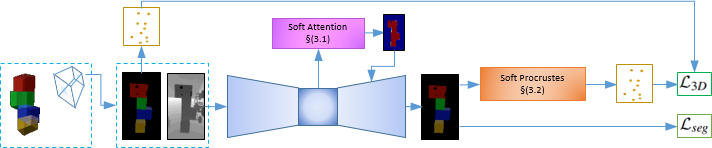}
\end{center}
   \vspace{-0.2in}
   \caption{We enhance a semantic segmentation model with a soft Procrustes analysis module which relies on a differentiable soft correspondence estimation.
   This bridges a dense classification objective with a regression one, penalizing misalignment between extracted keypoints $C$ from each labeled box side, and their corresponding box side centers $S$ of the virtual structure.
   This improves markerless volumetric sensor alignment by extending the space sensors can be successfully aligned in, allowing for more complex placements with minimal human intervention as correspondences are estimated without any markers, purely from each viewpoint's depth map with no markers required. \textbf{Top:} Multi-view concept. \textbf{Bottom:} Pipeline of model's training procedure. }
\label{fig:correspondences}
\vspace{-0.1in}
\end{figure}

Nevertheless, a low-cost system is not necessarily easy-to-use or portable.
The most commonly employed approach for multi-sensor spatial alignment involves moving a checkerboard within the overlapping field of views of adjacent cameras~\cite{beck15, Holoportation}.
It has the advantage of direct pairwise pose estimation and simultaneous intrinsic parameter estimation.
However, for setups that consist of a high number of sensors, this method needs to be supported by bundle adjustment as any errors get accumulated along the sensors' pose chain.
Therefore, it only serves as an initial pose estimation and a multi-sensor sparse feature correspondence establishment.
Still, traditional checkerboard calibration is a cumbersome process that requires human intervention and technical knowledge to execute correctly in a timely manner in order to avoid obsolete or ineffective checkerboard placements.

For example, checkerboard sweeping \cite{beck17} has been shown to reduce the time taken and even improve the quality of the alignment by calculating finer grained correction factors.
Despite such gains, the external optical tracking system requirement significantly limits its applicability due to the increased cost and the lack of portability.
Another approach is to use pre-defined markers, such as \textit{e.g.}~ArUco markers introduced in \cite{garrido2014automatic}, and associate them with known or measured landmarks in order to estimate each sensor's initial pose \cite{livescan3d}.
Though, given the stationary nature of this approach, a dense, geometry-based, global optimization is necessary to ensure correct alignment.
While this ensures minimum human intervention, it still requires technical knowledge regarding the marker positions and landmark establishment.

On the other hand, the Octolith structure introduced in \cite{Collet2015} poses as an alternative that minimizes both human intervention and technical knowledge, yet limiting portability and requiring specialized assembly. 
However, the use of a stationary anchor object is a very powerful alternative if combined with low-cost and easy-to-transport materials.
Motivated by this, Kowalski et al. \cite{livescan3d} as well as Alexiadis et al. \cite{alexiadis2017} design systems that rely on consumer-grade boxes that serve as anchor objects, positioned either randomly or in a pre-defined manner.
Both approaches estimated each sensor's initial pose with respect to the observed scene's known geometry, which is manually set in \cite{livescan3d}, and implicitly established from the virtual structure in \cite{alexiadis2017}, facilitated by marker detection.

Besides marker-based alignment methods, there are recent structure-based approaches that solely exploit the prior knowledge of the structure's geometry, eliminating the need for any visual markers like \cite{garrido2014automatic}, and thus, being truly markerless. In those cases, a single multi-view capture of the structure, which is placed arbitrarily inside the cameras' capture space, is sufficient for sensor pose estimation. In particular, this technique resulted in methods of multi-sensor volumetric alignment \cite{wscg2018, sterzentsenko2018low}, driven by a segmentation model trained using the known structure object. Such an approach facilitates ease of use, requires minimum human intervention and has no requirements for technical knowledge. However, it has the downside that a virtual 3D model of the structure's geometry must be available. On the other hand, this shortcoming is not severe for simple geometric structures, whose geometry can be trivially authored in a 3D modelling tool, such the ones used in \cite{wscg2018, sterzentsenko2018low} and in this paper.

A significant drawback of the previously mentioned approaches is that the segmentation models were trained on very limited camera poses and/or specific number of sensors in the multi-view capturing setup to ensure the robustness of the predictions.
Thus, albeit being practical and portable, they are limited by their flexibility in terms of placements and sensor count.

In this work, we surpass the aforementioned limitations by incorporating geometric constrains into a standard segmentation-based model, bridging the intermediate dense classification task with the targeted pose estimation one. This is achieved by introducing a novel, differentiable error term that regularizes the segmentation predictions of a deep autoencoder, and leads to better extrinsic calibration performance in expanded sensor placement configurations.

More specifically, the main contributions of the paper are:
\vspace{-0.05in}
\begin{itemize}
    \item A novel geometric objective, which is introduced to further optimize a deep convolutional autoencoder to estimate the pose of the calibration structure, after the initial semantic segmentation step. This objective is defined as the distance between the predicted keypoints and their point correspondences in the global coordinate system of the structure, and is depicted in Fig.~\ref{fig:correspondences}. Note, that the introduced error term is a result of a fully differentiable variant of the Procrustes analysis.
    \vspace{-0.05in}
    \item The proposed approach is not restricted by the number of sensors used for the spatial alignment, as the combination of the semantic segmentation and pose estimation tasks enables the independent processing of each sensor's output depthmap.
    \vspace{-0.05in}
    \item A soft attention mechanism is proposed, which forces the network to implicitly localize the calibration structure. This mechanism accelerates the convergence of the training process, while achieving robustness to pose variations.
\end{itemize}

%% file: RelatedWork.tex
\begin{figure*}[!hbtp]
\begin{center}
   \includegraphics[width=0.9\linewidth]{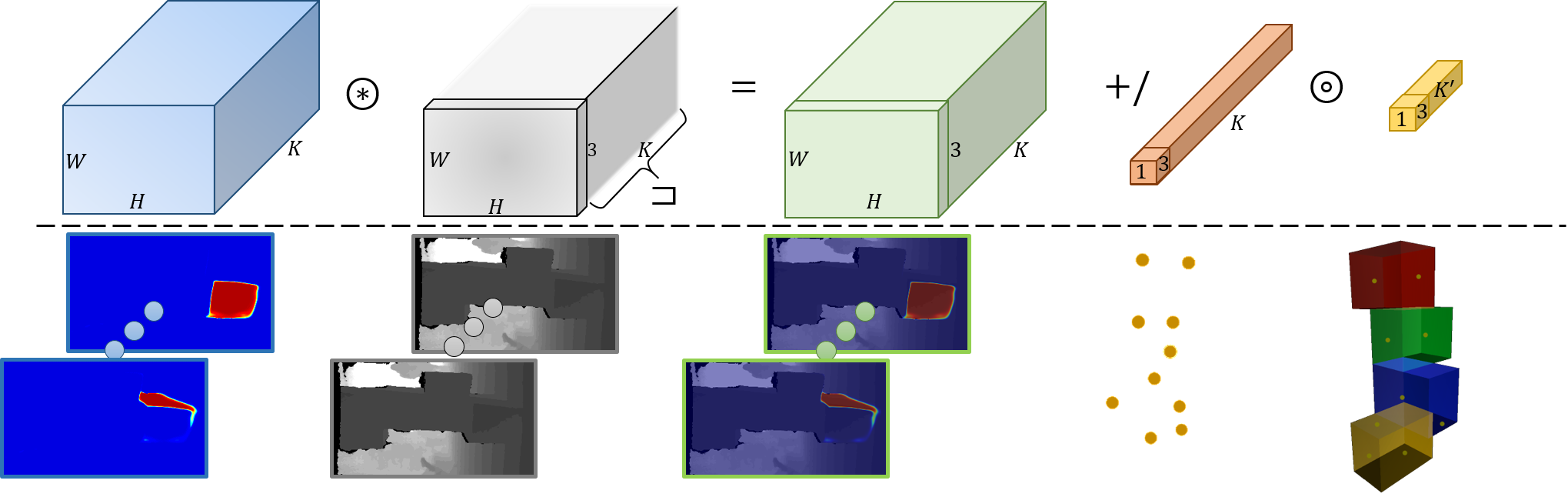}
\end{center}
   \vspace{-0.2in}
   \caption{The deep soft Procrustes analysis enables end-to-end geometric supervision for a semantic segmentation model.
   On the first row, the corresponding tensor operations are depicted. 
   Starting from a \textcolor{cyan}{light blue} $W \times H \times K$ tensor $P$ containing each of the $K$ classes' probabilities and the \textcolor{gray}{gray} $3 \times W \times H$ vertices tensor $V$ obtained by de-projecting the input depthmap,
   we establish soft correspondences as follows: \textbf{i)} we multiply ($\oast$) the tensors $P$ and $V$ after expanding ($\sqsupset$) -- or otherwise, broadcasting -- $V$ to $3 \times W \times H \times K$;
   \textbf{ii)} the resulting $3 \times W \times H \times K$ \textcolor{green}{light green} tensor $P \oast (V\sqsupset)$ is reduced via a mean operation across the spatial dimensions $W$ and $H$, resulting to the \textcolor{orange}{orange} $3 \times K$ tensor $C$ containing the soft correspondences' 3D coordinates; 
   \textbf{iii)} after masking with the ground truth labels and performing a SVD operation ($\ocirc$), the remaining correspondences in the \textcolor{yellow}{yellow} tensor $C^{\prime}$ are now aligned and any error function between them can be back-propagated to the semantic segmentation network. 
   The bottom row illustrates each operation's results visualizations.
   }
\label{fig:soft_corr_tensors}
\vspace{-0.2in}
\end{figure*}

Multi-view capture systems were pioneered in \cite{kanade1997virtualized} within the concept of \textit{Virtualized Reality}.
A dome of 51 cameras was built to capture users in 3D and replay their performances in free viewpoint rendering settings.
Their SoA evolution is the Panoptic Studio \cite{joo2017panoptic} a truncated pentagonal hexecontahedron structure with 480 low resolution cameras, 31 high resolution ones and 10 RGB-D sensors. 
The heterogeneous sensors' spatial alignment is achieved by projecting random patterns on a tent that is moved within the structure, with Structure-from-Motion aligning its calibration shot, and a subsequent bundle adjustment step consolidating the results of all shots into a single result.

Another SoA 3D capture system for high quality free viewpoint video is introduced by Collet~\textit{et al.}~\cite{Collet2015}.
In particular, 106 cameras, \textit{i.e.}~a mix between infrared and color ones, which are mounted on wheeled towers are spatially aligned using a customized octagonal tower structure called Octolith.
This is (re-)positioned and captured within the volumetric capture area five times to simultaneously calibrate the intrinsic and extrinsic parameters of the cameras.
Given that the Octolith has multiple checkerboard patterns on its faces, pairwise poses between the cameras are initially estimated and then refined using bundle adjustment.
While the former system relies on green chroma key background and professional studio lights, the Light Stage system, described in \cite{guo2019relightables}, employs 331 programmable lights in addition to the 90 infrared and color cameras mounted on a custom spherical dome.
All these sensors are spatially aligned using a traditional checkerboard process with Calibu circle markers establishing correspondences between cameras sub-groups which are then used to solve for a global pose solution using bundle adjustment.

Apart from these professional, expensive and non-portable volumetric capture setups, the commoditization of sensors in addition to the need for flexibility and portability has stimulated the development of lower-cost  solutions.
Checkerboard approaches are still heavily used in low sensor count systems. For example, in \cite{beck15} and \cite{beck17} the authors exploit the availability of external optical tracking to volumetrically align multiple RGB-D sensors.
While the former work \cite{beck15} uses static checkerboard sampling, the latter one \cite{beck17} moved towards sweeping checkerboard sampling to increase the usability of the process.
These methods leverage the depth information to additionally output dense look-up tables to correct the geometrical inputs.
This has also been exploited in a more coarse manner by Deng \textit{et al.}~\cite{deng2014registration} that estimate multiple local rigid transforms in the calibration volume to force depth measurement correction.
One of the major drawbacks of the checkerboard multi-sensor calibration is the reliance on a planar surface, which requires from the users to densely sample the area multiple times, establishing correspondences in sensor subgroups. 
As this is a cumbersome process, that requires additional technical knowledge with respect to the checkerboard's positioning, there have been efforts to reduce the time required (\textit{i.e.}~less samples).
In \cite{zhao2018marker} a set of ArUco \cite{garrido2014automatic} markers are attached on the sensors with an additional external one imaging both the checkerboard and the sensor attached markers.
In this way, less samples are required as more constraints are added into the optimization problem, reducing the time taken to calibrate the sensors and the chances of errors and/or problematic checkerboard placements. 
More recently, an automatic guidance system was developed by \cite{rojtberg2018efficient} and \cite{peng2019calibration} to interactively guide the user towards taking optimal samples.

Aiming to optimize the volumetric alignment process, the 8 view performance capture system of \cite{starck2007surface}, utilized a wand as the calibration object.
The advantage is that a symmetric object moved within the scene is visible by more or even all the sensors, effectively reducing the time taken to establish multi-view correspondences.
Apart from wand-like objects, fully symmetric spheres have also been used \cite{su2018fast} that are detected into the scene and either 2D (for simply cameras) or 3D (for depth sensors) correspondences are established for all views and then further optimized to estimate a global solution.
More innovative systems \cite{duncan2019fast} used the tracking offered by Virtual Reality (VR) controllers and rigidly attach detectable spheres on them in order to spatially align multiple sensors and simultaneously achieve alignment with the VR head-mounted display.

Even though the process itself is improved in terms of efficiency and ease of use by moving beyond the traditionally used planar objects, the fact that a user needs to manually operate the volumetric alignment process introduces management difficulties and hinders the process.
As a result, even earlier systems started utilizing structures positioned at the center of the captured volumes.
OmniKinect \cite{kainz2012omnikinect} attached markers on the faces of a custom calibration target and this is also the case for LiveScan3D \cite{livescan3d}.
The main difference of these methods is that for the former, the system is aligned with respect to a single camera, while for the latter, the users are required to input the markers' center 3D coordinates to achieve alignment on a common, \textit{i.e.} global, coordinate system.
Further, LiveScan3D then performs a dense 3D iterative closest point optimization step sequentially for each view's point cloud with respect to all other views' point clouds to refine the estimated solution.

Structure-based sensor alignments also use denser patterns, such as checkerboards, instead of markers in order to concurrently calibrate each sensor's intrinsics parameters with a prominent case being the Octolith used in \cite{Collet2015}.
A checkerboard cube structure was used in \cite{huang2019research} aligning each sensor with a specific face, with similar concepts used in \cite{tabb2019calibration} and \cite{ha2017deltille}.
Both these approaches enhance the traditional checkerboard with distinct ArUco markers on each side \cite{tabb2019calibration} or Deltille grids \cite{ha2017deltille} which are shown to improve calibration accuracy and can also be arranged in an icosahedron structure.

A recurring theme for all the aforementioned methods is their two-step nature.
Initially a set of features are extracted, either on the 2D image plane or directly as 3D coordinates.
These are estimated in relation to other sensors/viewpoints or fixed anchors (\textit{i.e.}~structures).
In this initial step, it is also possible to estimate an initial alignment of all sensors.
Then, following this initial sensor pose estimation, a subsequent dense optimization step offers a more refined and/or global solution for all sensors simultaneously.
This is done either in a pairwise manner \cite{livescan3d}, through Levenberg-Marquardt \cite{marquardt1963algorithm} or graph-based optimization \cite{owens2015msg, chen2019heterogeneous}.
This is estimated either with respect to a sensor or the anchor structure that defines the global coordinate system.
Evidently, it is the initial alignment step that estimates the correspondences and a preliminary pose for each sensor that needs to be optimized in terms of usability and practicality as this usually requires user intervention.
Users are required to either move objects within the capturing volume or position markers on a structure and the structure itself.
Our work improves upon recent works on depth-based volumetric alignment \cite{wscg2018, sterzentsenko2018low} that simplify this process by turning to pattern- and marker-less initial correspondence and sensor pose estimation.
They exploit the structure's geometry to densely annotate planar regions and extract correspondences at their centroids to estimate the initial pose.
The advantage of these methods is that they operate purely on the depth information, alleviating any issues related to pattern/marker detection due to illumination conditions.
Most of the aforementioned approaches rely on features acquired by the color images of RGB-D sensors to align them.
Apart from detection issues this is prone to color-to-depth misalignment.

Since our approach relies on semantic-driven soft Procrustes analysis, our work is also related to the recent advances in 6DOF pose estimation of known objects.
While preliminary approaches approached the problem directly by regressing the 6DOF pose, more recent approaches have managed to produce higher quality results by regressing 3D coordinates or keypoints instead.
In \cite{pavlakos20176} the heatmaps of semantic keypoints are regressed on the image with the final pose estimated by PnP, while in \cite{suwajanakorn2018discovery} these are automatically learned during training.
PVNet \cite{peng2019pvnet} densely regresses vectors pointing at the keypoints to improve robustness to occlusions.
More recent approaches regress object 3D coordinates at each pixel \cite{wang2019normalized, park2019pix2pose} in a normalized space to then fit the pose of the objects using their 3D representations.
Finally, PVN3D \cite{he2019pvn3d} relies on deep Hough voting to regress 3D keypoints directly and then fits the pose through least squares optimization.
In this work, we bridge the task of semantic segmentation of an object and that of pose estimation by adding geometric constraints during training of the segmentation network.
Essentially, each segment corresponds to a keypoint, which resembles the way this problem is approached in the literature currently by densely regressing per pixel attributes to allow for the localisation of keypoints. 
This improves the model's performance in larger 6DOF search spaces and allows us to estimate each sensor's initial pose to then drive the subsequent global optimization.

%% file: Approach.tex
In this section we outline our approach in more detail, starting with the basic principles of operation of our markerless volumetric sensor alignment method, which relies on densely extracted semantic correspondences as well as details about the model's architecture and its supervision scheme.
We then describe the integration of a geometric loss that accompanies the explicit dense classification objective with an implicit pose estimation objective in an end-to-end manner. 
This loss is based on a soft correspondence establishment technique through a differentiable Procrustes analysis.

%% file: SemanticCorrespondences.tex
By exploiting the principles introduced in \cite{wscg2018} and \cite{sterzentsenko2018low}, in this work we take a two step approach to solve the task of markerless spatial depth sensor alignment.
In the first step, assuming an a-priori known physical geometric structure, a global coordinate system is defined anchored at its virtual 3D model.
Given a single-view depth capture of this structure, we estimate the 3D coordinates of the structure's keypoints in the sensed data and establish 3D-point correspondences with the structure's virtual model.
Subsequently, we use those keypoint correspondences to perform sensor pose estimation with respect to the global coordinate system.
As a second final step that completes the volumetric alignment of multiple sensors, we perform a dense optimization refinement, using each sensor's initial pose estimate, reaching a global solution, as in \cite{sterzentsenko2018low}.

The geometric structure that we use here, is the same as the one presented in \cite{sterzentsenko2018low} and is a simple structure assembled from commercially available standardized packaging boxes. 
The idea behind the first step of our approach, is to train a deep convolutional autoencoder, which given a depthmap that represents an arbitrary view-point of the structure, will perform pixel-wise semantic segmentation in order to identify the visible planar sides of each box. 
The aforementioned semantic segmentation process facilitates keypoint extraction as the keypoints are placed in the mid-point of the box's planar sides. 
More specifically, a representative keypoint can be computed by averaging the 3D coordinates of the pixels belonging to each label. 
Under a correct depth-map labeling and sufficient side's visibility, this keypoint has a unique correspondence with the structure's virtual model, \textit{i.e}~the center of the respective box's side.

We use an adapted U-Net autoencoder~\cite{ronn}, depicted\footnote{A more detailed version can be found in the supplement.} in Fig.~\ref{fig:net}, comprising an encoder, a bottleneck, and a decoder. 
The encoder consists of 14 convolutional layers (CONV) each one followed by ReLU activations and downsamples the input depthmap 4 times using max pooling operators (POOL). 
Its output is fed into the bottleneck, which consists of 4 pre-activated ~\cite{he} residual blocks, each following a ReLU-CONV-ReLU-CONV structure. 
The decoder shares similar structure with the encoder, using 14 CONV followed by ReLU non-linearities. 
Note that the input feature map is upsampled 4 times prior to the segmentation prediction, using nearest neighbor interpolation. 
A $softmax$ function follows, which is applied at each pixel of the decoder output and serves as the initial estimation of the visible sides of each box. 

We train our model by jointly rendering synthetic views and label maps of the virtual model in a variety of poses. 
Note that prior work focus solely on the semantic segmentation task while in this work, we introduce a novel fully differentiable error term in the network's loss function which is based on Procrustes analysis. 
In particular, we minimize a total loss that is realised as:
\begin{equation}
    \pazocal{L}_{total} = \pazocal{L}_{seg} + \lambda \pazocal{L}_{3D},
\end{equation}
where $\pazocal{L}_{seg}$ is the per pixel negative log-likelihood of the predicted and ground truth planar visible side labels, normalized over the total number of pixels, $\pazocal{L}_{3D}$ is the geometrically derived objective that will be followingly defined in Eq.~\ref{eq:loss_function}, and $\lambda$ is a regularization term that controls the contribution of $\pazocal{L}_{3D}$ to the total loss.

In order to achieve faster convergence and improve the performance of our autoencoder, we introduce a soft-attention mechanism that forces the model to implicitly focus at the calibration structure. 
The mechanism that can be visualized in Fig.~\ref{fig:net}, is placed between the bottleneck part and the decoder of the model. 
In particular, given the activation matrix of the last residual block ${A} \in \mathbb{R}^{d\times h\times w}$, where $d$ is the number of channels of the activation map, we use a separate branch to convolve $A$ with a kernel of $1\times 1$ size, and apply a $sigmoid$ function to normalize the activation values to the $[0,1]$ space, thus forming an excitation mask $M \in [0,1]^{h\times w}$.
Note, that $M$ is element-wise multiplied with $A$ and then upsampled and re-applied to the activation maps after each upsampling layer of the decoder.

%% file: SoftProcrustes.tex
\begin{figure*}[!htbp]
\begin{center}
   \includegraphics[width=0.9\textwidth]{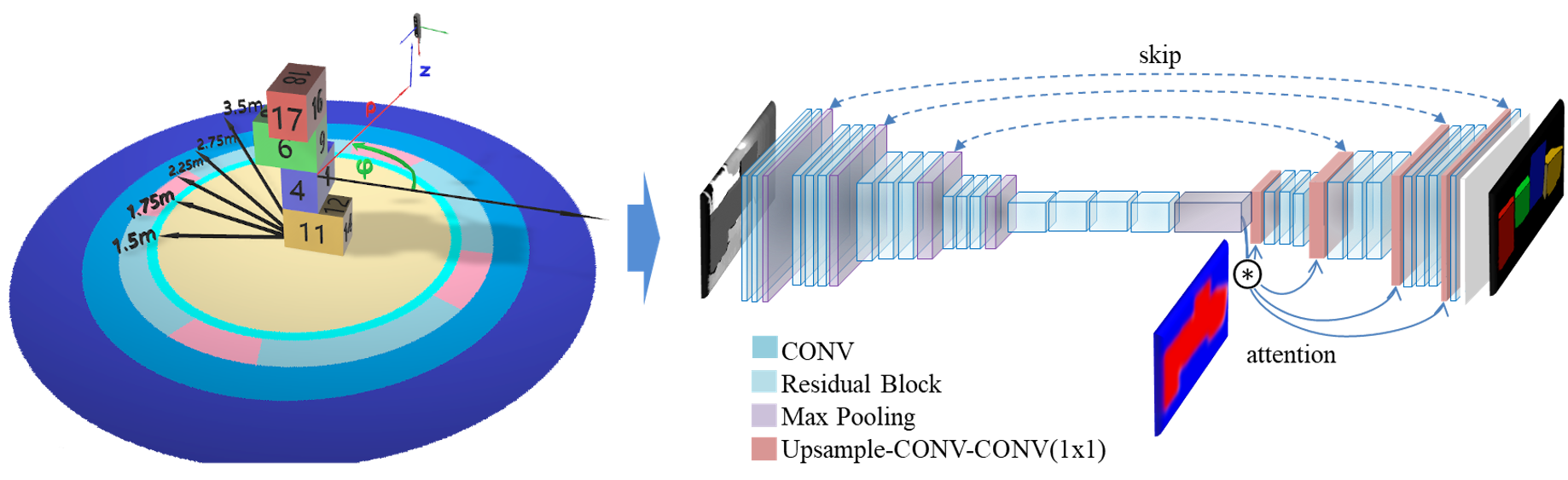}
\end{center}
   \vspace{-0.2in}
   \caption{Detailed representation of the capturing space and the model architecture of the proposed deep convolutional autoencoder. The model receives a raw depthmap as input and predicts per pixel segmentation labels. The soft attention mechanism, depicted as ``$\ast$" computes a heatmap, which is used to mask the activations of the last Residual Block, as well as the ones after each upsampling layer.}
\label{fig:net}
\vspace{-0.2in}
\end{figure*}

In order to enhance training by infusing an end task specific goal, we introduce a geometric objective to optimize, which in this case is the initial pose estimation through a Procrustes analysis.
However, this requires correspondence estimation that relies on the dense classification result, which is usually estimated by the non-differentiable $argmax$ operation.
To facilitate end-to-end training, we introduce a soft correspondence estimation establishment within the network that allows for gradient back-propagation.

Let $P \in [0,1]^{K \times H \times W}$, with $\sum_{k} P_{k,h,w} = 1, \forall k \in \{0, 1, ..., K-1\}, (h,w) \in \{0,1,..., H-1\} \times \{0, 1, ..., W-1\}$ be a probability map, with $P_{k,h,w}$ denoting the probability of pixel $(h,w)$ belonging to class $k$, as predicted by the network. 
This is the output of the segmentation network predicting per pixel probabilities for $K$ classes, after the $softmax$ operation.
Let also $V \in \mathbb{R}^{3 \times H \times W}$ denote the 3D coordinates of the de-projected depth-map and $C \in \mathbb{R}^{3 \times K}$ denote the extracted keypoints of each labeled box side. 
Then, we can define our soft correspondence extraction as the weighted average across each class's dense probability map:
\begin{equation}
\label{eq:soft_corr}
    C_{i,k} = \frac{\sum_{h,w} P_{k,h,w} V_{i,h,w}}{\sum_{h,w}P_{k,h,w}}.
\end{equation}
Let $S \in \mathbb{R}^{3 \times K}$ denote the 3D coordinates of the $K$ box side centers.
In the case of perfectly correct semantic segmentation and provided that the respective structure's box sides are visible, the keypoints computed by Eq.~(\ref{eq:soft_corr}) can be perfectly aligned with the global coordinate system using Procrustes analysis after establishing the correspondences $C \leftrightarrow S$. 
This soft correspondence estimation, depicted in Fig.~\ref{fig:soft_corr_tensors}, enables the integration of a geometric pose estimation objective into the network during training.

This will supplement semantic segmentation through a 3D keypoint correspondence error term:
\begin{align}
\label{eq:loss_function}
    \pazocal{L}_{3D} & = ||M \cdot (\hat{S}-\Omega\hat{C})||_F ,
\end{align}
where ``$\cdot$" denotes the Hadamard product, $||\cdot||_F$ the matrix Frobenius norm, $M = \{0,1\}^{3\times K}$ the ground truth pixel visibility mask, and $\hat{S},\hat{C}$ the corresponding $S,C$ normalized using the average values computed across their second dimension (\textit{e.g.}~$\hat{S} = S - \bar{S}_{K}$). $\Omega$ represents the solution to the orthogonal Procrustes problem of aligning $\hat{S}$ with $\hat{C}$ via a rotation matrix $\Omega$ obtained by the Singular Value Decomposition (SVD).

%% file: Results.tex
In this section we provide the evaluation strategy that we followed in order to assess the effectiveness of the proposed method. 
We begin by discussing the implementation details that apply globally to our evaluation strategy. 
Subsequently, we split our evaluation in two parts. 
Initially, in Section \ref{sec:self_evaluation} we compare the performance of the proposed soft Procrustes and soft attention methods against equivalent CNN networks that have those mechanisms removed, \textit{i.e.}~assessing the benefits of adopting those mechanisms under the same overall CNN architecture. 
Then, taking into account this analysis and using the best performing model of our proposed method, in Section \ref{sec:comparisson_other_methods} we compare the performance of this model in the end-to-end task of extrinsic depth camera calibration in multi-view setups against SoA marker-based and markerless methods.

\textbf{Implementation details.} Similarly to \cite{sterzentsenko2018low}, the proposed CNN is trained exclusively using a synthetic dataset. We assume a global coordinate system (GCS) with respect to the virtual 3D model of the structure, with its origin located at the center of the structure, the $y$-axis going up towards the sky, while axii $x$ and $z$ extending in parallel to the floor. Further, we define a parametric space of potential camera positions in cylindrical coordinates ($\rho$,$\phi$,$z$) with the parametric space's respective unit vectors $\mathbf{\dot{\rho}}$ and $\mathbf{\dot{\phi}}$ being in parallel to the global $x$-$z$ plane and $\mathbf{\dot{z}}$ being aligned with the GCS's $y$ axis. For the camera orientation, we define the camera's local coordinate system to have $z$ axis pointing to the looking direction, $y$ axis going down and $x$ axis going right (Please refer to Fig. \ref{fig:net} for a visualization of the coordinate systems). When synthetically generating a camera pose, we first sample uniformly its position from a predefined range of $\rho$, $\phi$ and $z$. Then, for its orientation, we define a potential look at radius range $r_l$ around GCS's origin and randomly generate a look direction inside this radius again by uniform sampling. For all the poses that we generate, we apply a random rotation around the camera's $z$ axis up to $t$ degrees after aligning the camera's $x$ axis to be parallel to the GCS's $x$-$z$ plane. In all of our experiments we use $\phi \in [0^\circ, 360^ \circ]$, $z \in [-15,75]$cm (which corresponds to a height from the ground $[70-160]$cm, sufficient for most multi-view camera setups), $l_r = 20$cm and $t = 5^\circ$. For each generated sample, the projection matrix that we used for rendering the virtual model was randomly sampled from a pool of actual depth sensor camera intrinsic matrices.

Given a synthetically generated viewpoint, \textit{i.e.}~camera pose, we render depth and label maps of the structure. Our structure that consists of four orthogonal parallelepiped packaging boxes, has $24$ box sides in total. Since the floor facing sides of the boxes are hardly ever seen by any camera, we exclude them from labeling, and train our CNN with $21$ classes, accounting for the remaining box sides, plus one class for pixels located in the background. 

In order for our model to better generalize to input depthmaps captured by actual sensors and further increase its robustness, we introduced a depthmap noise augmentation policy, which simulates, in a simple way, common noise patterns that appear in depthmaps captured by commercial-grade depth sensors. In particular, the synthetic depth-maps are augmented by introducing zero-depth values to the rendered structure's borders as well as artificially introducing holes, to account for regions of invalid or no measurement appeared in real captured images depicting low to none textured objects (stereo technology) or objects that consist of absorptive/reflective materials (time of flight technology).

Regarding the training of the CNN, we choose to initialize the weights of the proposed deep convolutional autoencoder using Xavier initialization~\cite{glorot2010understanding}. Further, we use Adam optimization~\cite{adam} with $\beta1 = 0.9$. $\beta2= 0.99$. The learning rate is set to $0.0002$, while using $\lambda=0.1$ and a mini-batch size of 16. The training process converges after approximately 3200K iterations. The model was implemented using the PyTorch framework~\cite{pytorch}, while each model was trained on one NVIDIA GeForce GTX 1080 graphics card, with each train lasting approximately 4 days each.

\subsection{Soft Procrustes and Soft Attention evaluation}
\label{sec:self_evaluation}

\begin{figure*}[!htbp]
\begin{center}
   \includegraphics[width=\linewidth]{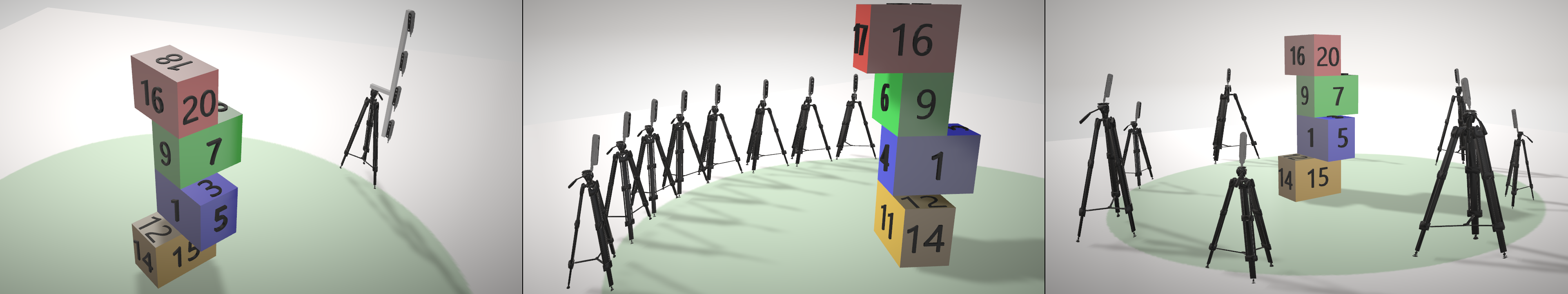}
\end{center}
   \vspace{-0.2in}
   \caption{Virtual representations of setups on which methods were evaluated. In more detail, please refer to section \ref{sec:results}. From left to right: (\textbf{a}) shows the \textit{array configuration} used to capture real data for comparison as described in \ref{sec:self_evaluation}; (\textbf{b}) showcases the \textit{arc configuration}; and (\textbf{c}) shows a full 360 setup with alternating camera heights. 
   Both (\textbf{b}) and (\textbf{c}) used for calibration setup evaluation and comparison against other methods, further described in  \ref{sec:comparisson_other_methods}.}
\label{fig:test_samples}
\vspace{-0.2in}
\end{figure*}

In this section we evaluate 7 network models based on the CNN architecture described in Section \ref{sec:approach} and perform an ablation study with respect to soft Attention and soft Procrustes mechanisms. We trained the models in 3 different ranges for parameter $\rho$ in order to assess the effectiveness of the proposed method with respect to the size of the parametric camera pose space. Table \ref{tab:models_description} summarizes the characteristics of each trained model.

\begin{table}[]
\caption{Definition of the evaluated models: ``S-Att" indicates that the model utilizes the soft attention mechanism, ``S-Proc" indicates the utilization of the soft Procrustes mechanism, while $\rho$ is sensor-to-structure range.}
\label{tab:models_description}
\vspace{-0.2in}
\small
\begin{center}
\begin{tabular}{c c c c}
\hline
\textbf{Model} & \textbf{S-Att} & \textbf{S-Proc} & \textbf{$\rho$ (cm)} \\ \hline
m01 & \xmark & \xmark & 150-225 \\
m02 & \cmark & \xmark & 150-225 \\
m03 & \cmark & \cmark & 150-225 \\ \hline
m04 & \cmark & \xmark & 150-275 \\
m05 & \cmark & \cmark & 150-275 \\ \hline
m06 & \cmark & \xmark & 150-350 \\
m07 & \cmark & \cmark & 150-350 \\ \hline
\end{tabular}
\end{center}
\vspace{-0.2in}
\end{table}

\begin{table}[]
\caption{Ablation study in the dataset with $\rho = 165$cm. Mean, Standard Deviation (STD) and Relative Standard Deviation (RSD) for each metric are reported. The last column of each metric reports performance decrease wrt the average performance of the best model. mIoU is expressed in percentage while 3D RMSE is expressed in meters}
\label{tab:Dataset165}
\vspace{-0.3in}
\begin{center}
\scalebox{0.65}{
\begin{tabular}{l|llll|l|l|l|l}
\hline
\multicolumn{9}{c}{\textbf{Dataset 165cm}} \\ \hline
 & \multicolumn{4}{c}{\textbf{mIoU}} & \multicolumn{4}{c}{\textbf{3D RMSE}} \\ \hline
\multicolumn{1}{c|}{\textbf{Model}} & \multicolumn{1}{c|}{\textbf{Mean}} & \multicolumn{1}{c|}{\textbf{STD}} & \multicolumn{1}{c|}{\textbf{RSD}} & \multicolumn{1}{c|}{\textbf{}} & \multicolumn{1}{c|}{\textbf{Mean}} & \multicolumn{1}{c|}{\textbf{STD}} & \multicolumn{1}{c|}{\textbf{RSD}} & \multicolumn{1}{c}{} \\ \hline
m01 & \multicolumn{1}{l|}{\textbf{89.47\%}} & \multicolumn{1}{l|}{7.78\%} & \multicolumn{1}{l|}{8.7\%} & \textbf{+0.00\%} & 0.1050 & 0.0991 & 94.4\% & +17.98\% \\ 
m05 & \multicolumn{1}{l|}{89.43\%} & \multicolumn{1}{l|}{\textbf{5.57\%}} & \multicolumn{1}{l|}{\textbf{6.2\%}} & -0.05\% & \textbf{0.0890} & \textbf{0.0431} & \textbf{48.4\%} & \textbf{+0.00\%} \\ 
m04 & \multicolumn{1}{l|}{89.17\%} & \multicolumn{1}{l|}{6.78\%} & \multicolumn{1}{l|}{7.6\%} & -0.34\% & 0.0924 & 0.0587 & 63.5\% & +3.82\% \\ 
m02 & \multicolumn{1}{l|}{89.08\%} & \multicolumn{1}{l|}{6.97\%} & \multicolumn{1}{l|}{7.8\%} & -0.45\% & 0.0980 & 0.0552 & 56.3\% & +10.13\% \\ 
m03 & \multicolumn{1}{l|}{88.51\%} & \multicolumn{1}{l|}{6.91\%} & \multicolumn{1}{l|}{7.8\%} & -1.08\% & 0.1270 & 0.1207 & 95.0\% & +42.62\% \\ 
m07 & \multicolumn{1}{l|}{87.58\%} & \multicolumn{1}{l|}{8.54\%} & \multicolumn{1}{l|}{9.7\%} & -2.12\% & 0.1192 & 0.1124 & 94.3\% & +33.94\% \\ 
m06 & \multicolumn{1}{l|}{87.22\%} & \multicolumn{1}{l|}{8.46\%} & \multicolumn{1}{l|}{9.7\%} & -2.52\% & 0.1155 & 0.1439 & 124.6\% & +29.73\% \\ \hline
\end{tabular}}
\end{center}
\vspace{-0.2in}
\end{table}

\textbf{Datasets.} To assist our evaluation of the aforementioned models, we created datasets, with real multi-view captures of the proposed structure. Four (4) Intel RealSense D415 sensors were placed horizontally in a vertical array configuration with the in-between sensor distance to be approximately 25cm. The array configuration was later attached on a camera tripod as depicted in Fig.~\ref{fig:test_samples} (a). During the capturing process, the calibration structure was placed at $3$ distinct distances from the cameras, at a range $\rho$ of $165$cm, $200$cm and $235$cm, each one comprising a distinct dataset. With respect to the ground floor, the heights of the cameras were approximately placed at heights $80$cm, $105$cm, $130$cm and $155$cm. In order to avoid relocating the tripod with all the mounted devices and their attached computing units, \textit{i.e.}~Laptops and PCs, we rotated the structure around it's $y$ axis, simulating an all-around camera configuration. Specifically, the structure was rotated $8$ times, $\approx 45^\circ$ at a time, covering a full $360^\circ$ view of the calibration structure, resulting into $32$ views per distinct distance and $96$ views in total. 

\textbf{Metrics.} Two different metrics were used to evaluate the considered models. While, semantic segmentation tasks are usually evaluated using the mean intersection over union (mIoU) metric  between the predicted and ground truth labeled data, we take a slight modification of this approach that takes into account the 3D nature of the underlying pose estimation task. In particular, given the semantic segmentation predictions of the models, we extract the structure's keypoints, estimate the camera's pose via SVD and render the virtual model of the structure from the computed viewpoint and at last compute mIoU between rendered and predicted labels. In the case of a correct pose estimation, the rendered view's labeling with the predicted semantic segmentation of the network should be perfect, resulting in a high mIoU value. As a second metric to evaluate the performance of the proposed models we use the standard root mean squared error (RMSE) between the 3D euclidean distances of the visible points belonging to the aligned point-sets $C$ and $S$, after Procrustes analysis.

\textbf{Evaluation Methodology.} We evaluate our models (which are trained for different ranges of $\rho$), against our 3 datasets which correspond to different $\rho$ values. First, we perform a comparison against the Dataset with $\rho=165$cm. All models participate in this comparison, since this value of $\rho$ was included in the training set of all models. Then, we compare against the union of datasets with $\rho=165$cm and $\rho=200$cm. In this case, models m01, m02, m03 are excluded from the comparison, since their training set does not include the range $\rho=200$cm. Finally, we compare m04, m05, m06 and m07 against the union of all datasets (i.e $\rho=165$cm, $\rho=200$cm, $\rho=235$cm).
Thus, we do a systematic comparison by incrementally including camera placement configurations which correspond to increased $\rho$ distances. 

\textbf{Results for dataset 165cm.} Table~\ref{tab:Dataset165} presents the performance of all studied models with respect to the first dataset of $\rho = 165$cm. We begin our analysis with a pair-wise comparison of the models (m01, m02, and m03) trained in a camera pose parametric space closer to the dataset's $\rho$ distance. Our simplest model, m01, does not integrate any of the soft Attention or soft Procrustes mechanisms, however, with respect to the mIoU metric, achieves the best performance among all models. Despite its top performance on average, it has the worst performance robustness among all the other models we consider here, as implied by its worst absolute (and relative) standard deviation (STD / RSD). The next top performing model among the considered set, on the same metric, is m02, which integrates the proposed soft attention mechanism. While it demonstrates a performance decrease of $0.45\%$ in mIoU terms, it has a smaller STD than m01 and performs best among its immediate competitors with respect to the 3D RMSE metric in both average and STD terms. Our argument for low robustness of m01 is further supported by the fact that m01 has the worse STD/RSD in 3D RMSE terms from all the aforementioned models by a far margin. 

Conclusively, we found that applying the proposed soft attention mechanism the model's performance in the camera pose estimation task can be improved (as justified by the 3D RMSE metric), while increasing its robustness (as justified by its lower STD). Thus, in the experiments that follow, we integrated this soft attention mechanism to all our models. Comparing the model with the soft Procrustes mechanism (m03) with the previous models trained in the same set, we've found that it is the worst performing model among the previous ones. In that particular case, soft Procrustes did not offer any performance advantage over the other methods.

\begin{table}[!t]
\caption{Ablation study in the datasets with $\rho \in \{165cm, 200cm\}$. Mean, Standard Deviation (STD) and Relative Standard Deviation (RSD) for each metric are reported. The last column of each metric reports performance decrease wrt the average performance of the best model. mIoU is expressed in percentage while 3D RMSE is expressed in meters}
\label{tab:Dataset165200}
\vspace{-0.3in}
\begin{center}
\scalebox{0.7}{
\begin{tabular}{l|llll|l|l|l|l}
\hline
\multicolumn{9}{c}{\textbf{Dataset 165 \& 200 cm}} \\ \hline
 & \multicolumn{4}{c}{\textbf{mIoU}} & \multicolumn{4}{c}{\textbf{3D RMSE}} \\ \hline
\multicolumn{1}{c|}{\textbf{Model}} & \multicolumn{1}{c|}{\textbf{Mean}} & \multicolumn{1}{c|}{\textbf{STD}} & \multicolumn{1}{c|}{\textbf{RSD}} & \multicolumn{1}{c|}{\textbf{}} & \multicolumn{1}{c|}{\textbf{Mean}} & \multicolumn{1}{c|}{\textbf{STD}} & \multicolumn{1}{c|}{\textbf{RSD}} & \multicolumn{1}{c}{} \\ \hline
m05 & \multicolumn{1}{l|}{\textbf{89.90}\%} & \multicolumn{1}{l|}{\textbf{5.12}\%} & \multicolumn{1}{l|}{\textbf{5.7}\%} & \textbf{+0.00\%} & \textbf{0.0900} & \textbf{0.0400} & \textbf{44.4\%} & \textbf{+0.00\%} \\ 
m04 & \multicolumn{1}{l|}{89.70\%} & \multicolumn{1}{l|}{5.13\%} & \multicolumn{1}{l|}{5.7\%} & -0.22\% & 0.1019 & 0.0951 & 93.3\% & +13.27\% \\ 
m07 & \multicolumn{1}{l|}{88.37\%} & \multicolumn{1}{l|}{7.19\%} & \multicolumn{1}{l|}{8.1\%} & -1.70\% & 0.1091 & 0.1020 & 93.5\% & +21.26\% \\ 
m06 & \multicolumn{1}{l|}{88.01\%} & \multicolumn{1}{l|}{7.23\%} & \multicolumn{1}{l|}{8.2\%} & -2.11\% & 0.1082 & 0.0910 & 84.1\% & +20.26\% \\ \hline
\end{tabular}}
\end{center}
\vspace{-0.1in}
\end{table}

\begin{table}[]
\caption{Ablation study in the datasets with $\rho \in \{165cm, 200cm, 235cm\}$. Mean, Standard Deviation (STD) and Relative Standard Deviation (RSD) for each metric are reported. The last column of each metric reports performance decrease w.r.t. the average performance of the best model. mIoU is expressed in percentage while 3D RMSE is expressed in meters}
\label{tab:Dataset165200235}
\vspace{-0.3in}
\begin{center}
\scalebox{0.7}{
\begin{tabular}{l|llll|l|l|l|l}
\hline
\multicolumn{9}{c}{\textbf{Dataset 165 \& 200 \& 235cm}} \\ \hline
 & \multicolumn{4}{c}{\textbf{mIoU}} & \multicolumn{4}{c}{\textbf{3D RMSE}} \\ \hline
\multicolumn{1}{c|}{\textbf{Model}} & \multicolumn{1}{c|}{\textbf{Mean}} & \multicolumn{1}{c|}{\textbf{STD}} & \multicolumn{1}{c|}{\textbf{RSD}} & \multicolumn{1}{c|}{\textbf{}} & \multicolumn{1}{c|}{\textbf{Mean}} & \multicolumn{1}{c|}{\textbf{STD}} & \multicolumn{1}{c|}{\textbf{RSD}} & \multicolumn{1}{c}{} \\ \hline
m05 & \multicolumn{1}{l|}{\textbf{89.83\%}} & \multicolumn{1}{l|}{4.53\%} & \multicolumn{1}{l|}{\textbf{5.0\%}} & \textbf{+0.00\%} & \textbf{0.1028} & \textbf{0.0793} & 77.2\% & \textbf{+0.00\%} \\ 
m04 & \multicolumn{1}{l|}{89.65\%} & \multicolumn{1}{l|}{\textbf{4.49}\%} & \multicolumn{1}{l|}{\textbf{5.0\%}} & -0.20\% & 0.1124 & 0.0852 & 75.8\% & +9.37\% \\ 
m07 & \multicolumn{1}{l|}{88.56\%} & \multicolumn{1}{l|}{7.10\%} & \multicolumn{1}{l|}{8.0\%} & -1.42\% & 0.1141 & 0.0828 & 72.6\% & +11.02\% \\ 
m06 & \multicolumn{1}{l|}{87.82\%} & \multicolumn{1}{l|}{6.91\%} & \multicolumn{1}{l|}{7.9\%} & -2.24\% & 0.1192 & 0.0918 & 77.1\% & +15.98\% \\ \hline
\end{tabular}}
\end{center}
\vspace{-0.2in}
\end{table}

However, when we tried to enlarge the camera pose parametric space of the training set, by increasing the maximum $\rho$ value, we observed that soft Procrustes could actually offer a performance improvement. Comparing m05 with m04 (both trained in the same parametric space) we find that m05, \textit{i.e.}~the one with soft Procrustes, outperforms m04 in all metrics, both on average and STD terms.  A similar conclusion can be drawn when also comparing m07 with m06. Soft Procrustes has shown better mIoU performance than its competitor on average terms, close performance in 3D RMSE average terms and better performance in 3D RMSE STD/RSD terms.

Overall, among all the models that we trained, m05 with soft Procrustes performed best in this dataset in all 3D RMSE terms and marginally close to m01 in mIoU average terms, while being better in mIoU STD/RSD terms than every other model. 

\begin{figure}[!tp]
\begin{center}
   \includegraphics[width=0.95\columnwidth]{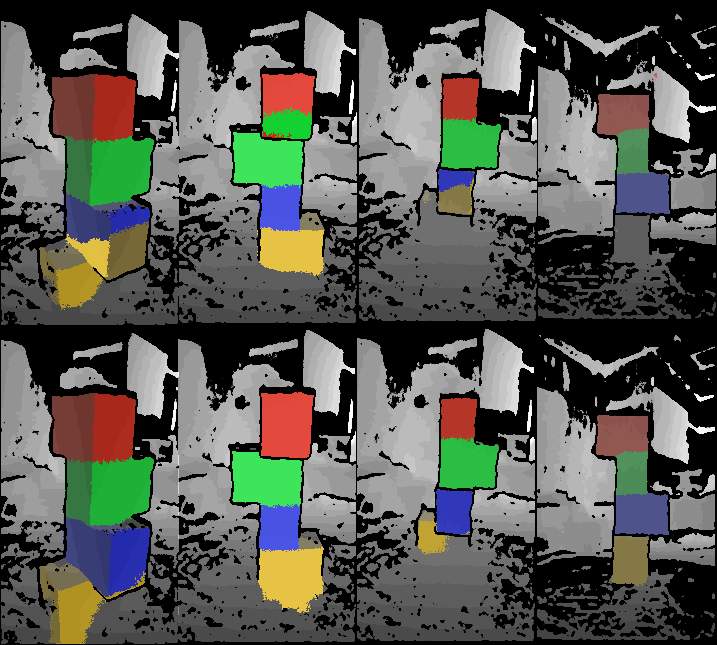}
\end{center}
   \vspace{-0.2in}
   \caption{ Qualitative comparison between m04 (top) and m05 (bottom) models, showcasing the gains of the soft Procruses model (m05), as it correctly identifies and labels the box sides in challenging poses where m04 fails. 
   However, m04 behaves better at the structure's faces close to the floor.}
\label{fig:segmentations}
\vspace{-0.3in}
\end{figure}

\textbf{Results for datasets 165 \& 200cm.} Table~\ref{tab:Dataset165200} presents the performance of the models that are trained in distances $\rho$ larger than 200cm, when evaluating them in the union of the datasets of 165cm and 200cm. Once again, in this experiment the soft Procrustes model, m05, showcases the best performance in mIoU and 3D RMSE terms, being by far the best performing model among the rest. The contribution of soft Procrustes is still visible but less apparent when comparing m07 to m06. The soft Procrustes model performs better in mIoU terms but not as good as m06 in RMSE terms.

\textbf{Results for datasets 165 \& 200 \& 235cm.} Finally, the results of our last experiment, conducted in the union of our datasets of $\rho = \{165,200,235\}$cm, are depicted in Table \ref{tab:Dataset165200235}. The m05 soft Procrustes model has shown the best performance across all datasets in mIoU and 3D RMSE terms. Furthermore, the soft Procrustes model m07 has better performance in 3D RMSE terms than m06. However, regarding mIoU, it shows worse performance in STD/RSD terms, even though on average it is still better than m06.

\textbf{Conclusion.} Following our previously presented experiments conducted in incrementally enlarged datasets of camera poses, we observe than the proposed soft Procrustes mechanism is able to enhance performance and robustness of the CNN model, with its contribution becoming more important, as the parametric space of the training set is enlarged. While in the comparisons of m05 with m04 the correctness of the proposed argument is strongly supported, as it is depicted in Fig.~\ref{fig:segmentations}, the experiments did not fully prove that m07 is better than m06 in all aspects. We may argue, that the reason for this might be that these two models have been trained in a much larger parametric space than the pose space covered by our real-world captured evaluation datasets. Our experiments support our thesis at least partially, since m07 has always been better than m06 with respect to at least one metric in all of our cases. Potential future experimentation may provide stronger proof of this statement.

\begin{figure}[!tp]
\begin{center}
   \includegraphics[width=0.95\columnwidth]{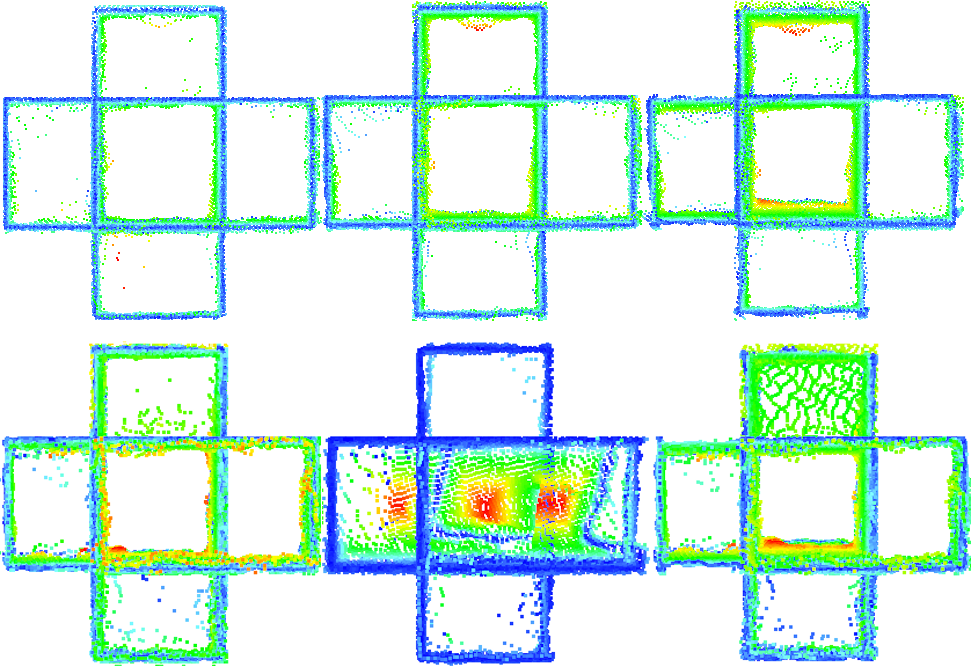}
\end{center}
   \vspace{-0.2in}
   \caption{Calibration results (from left to right): \textbf{a)} marker-based , \textbf{b)} \cite{sterzentsenko2018low} and \textbf{c)} proposed method, after global refinement. Each column corresponds to a different method and each row to a different test case.
   Note that warmer colors indicate large errors while colder ones indicate lower errors.
   \textbf{Row 1}: The markerless methods produce results very close to the marker-based in column 1.
   \textbf{Row 2}: The marker-based results are of lower quality, \cite{sterzentsenko2018low} failed to align correctly one of the viewpoints, resulting in an inaccurate calibration, while the soft Procrustes model successfully aligned all views with close accuracy to the marker-based method.}
\label{fig:four_top}
\vspace{-0.1in}
\end{figure}

\subsection{Evaluation over existing calibration methods} 
\label{sec:comparisson_other_methods}
In this section, the best performing model of our work (m05) is compared against two other structure-based methods on the camera extrinsics calibration task, namely \cite{sterzentsenko2018low} and a variant of \cite{livescan3d}. In particular, \cite{livescan3d} is a marker-based calibration method which we adapted to our specific calibration structure. Markers generated by \cite{garrido2014automatic}, are placed on every side of each box while ensuring that the center of the marker is aligned with the center of the respective box's side. Marker detection facilitates correspondence establishment between the detected marker's center and the centroid of the respective box's side in the structure's virtual model. In our experiments with the Intel RealSense D415 devices, apart from depth-maps, we had to also capture the sensors' Infrared (IR) stream, which is spatially aligned with the sensor's depth stream enabling easy estimation of the marker's 3D position in camera space. As in our case, standard Procrustes Analysis is subsequently used, in order to align the positions of the markers in camera space, with their corresponding 3D points in the structure's virtual 3D model and thus estimating the camera's pose with respect to the GCS anchored at the virtual model of the structure.

While our work extends the markerless method of \cite{sterzentsenko2018low}, a direct comparison with that method is limited, due to the fact that \cite{sterzentsenko2018low} can only calibrate a multi-view setup of exactly 4 cameras placed on a very specific configuration.

\begin{table}[!t]
\caption{Quantitative report on the \textit{arc} dataset across marker-based and proposed methods evaluated on Mean RMSE, Standard Deviation  and Relative Standard Deviation metrics.}
\label{tbl:DatasetArc}
\vspace{-0.1in}
\begin{center}
\scalebox{0.85}{
\begin{tabular}{l|l|l|l|l}
\hline
\multicolumn{5}{c}{\textbf{Dataset: arc}} \\ \hline
\multicolumn{1}{c|}{\textbf{Method}} & \multicolumn{1}{c|}{\textbf{Mean}} & \multicolumn{1}{c|}{\textbf{STD}} & \multicolumn{1}{c|}{\textbf{RSD}} & \multicolumn{1}{c}{} \\ \hline
markerbased\_refined & \textbf{0.0136} & \textbf{0.0028} & \textbf{3.76\%} & \textbf{0.00\%} \\
markerless\_refined & 0.0142 & 0.0034 & 4.91\% & +4.58\% \\ 
markerbased\_initial & 0.0378 & 0.0122 & 10.36\% & +178.57\% \\ 
markerless\_initial & 0.0579 & 0.0126 & 9.06\% & +326.37\% \\
\hline
\end{tabular}}
\end{center}
\vspace{-0.3in}
\end{table}

\begin{table*}[!ht]
\caption{Quantitative report on the \textit{cameras\_6\_8} (left half of the table) across marker-based and proposed methods,as well as on \textit{cameras\_4} (right half of the table) dataset across all methods, evaluated on Mean, Standard Deviation and Relative Standard Deviation metrics.}
\label{tbl:DatasetCameras468}
\vspace{-0.2in}
\begin{center}
\scalebox{0.9}{
\begin{tabular}{l|l|l|l|l|l|l|l|l}
\hline
\multicolumn{1}{c}{\textbf{}} & \multicolumn{4}{c}{\textbf{Dataset: cameras\_6\_8}} & \multicolumn{4}{c}{\textbf{Dataset: cameras\_4}} \\ \hline
\multicolumn{1}{c|}{\textbf{Method}} & \multicolumn{1}{c}{\textbf{Mean}} & \multicolumn{1}{c}{\textbf{STD}} & \multicolumn{1}{c}{\textbf{RSD}} & \multicolumn{1}{c}{} & \textbf{Mean} & \textbf{STD} & \textbf{RSD} &  \\ \hline
markerbased\_refined & 0.0273 & \textbf{0.0007} & 2.67\% & 3.93\% & \textbf{0.0288} & \textbf{0.0008} & \textbf{2.91\%} & \textbf{0.00\%} \\
markerless\_refined & \textbf{0.0262} & \textbf{0.0007} & \textbf{2.64\%} & \textbf{0.00\%} & 0.0292 & 0.0013 & 4.50\% & 1.42\% \\ 
\cite{sterzentsenko2018low} initial & \multicolumn{4}{c|}{N/A} & 0.0330 & 0.0011 & 3.38\% & 14.62\% \\ 
\cite{sterzentsenko2018low} & \multicolumn{4}{c|}{N/A} & 0.0338 & 0.0016 & 4.71\% & 17.18\% \\ 
markerbased\_initial & 0.0348 & 0.0042 & 12.16\% & 32.63\% & 0.0361 & 0.0072 & 19.90\% & 25.20\% \\ 
markerless\_initial & 0.0556 & 0.0027 & 4.87\% & 112.07\% & 0.0552 & 0.0069 & 12.51\% & 91.44\% \\ \hline
\end{tabular}}
\end{center}
\vspace{-0.3in}
\end{table*}

\textbf{Datasets.} For our evaluation purposes, we captured 3 multi-view datasets of the calibration structure in varying placement configurations that we used in order to compare our method with \cite{livescan3d} and \cite{sterzentsenko2018low}. For the first dataset (``arc''), we used 8 depth sensors in an arc configuration of $\approx 120^\circ$, as depicted in Fig.~\ref{fig:test_samples} (b). All the cameras were placed at two different height levels from the ground floor, namely 90cm and 110cm and at 3 different distances from the calibration structure, namely 165cm, 200cm and 235cm, all one at a time. At each camera placement configuration we made 4 captures of the structure, while rotating the structure around its $y$ axis for about $\approx 90^\circ$  in between each capture. This dataset contains 24 multi-view captures in total.

The second dataset (``cameras\_6\_8'') that we used for our evaluation, contains multi-view captures of the calibration structure, captured by two full $360^\circ$ camera setups of 6 and 8 cameras respectively, evenly partitioning the circular area around the structure in each case. All cameras in this dataset configurations were equally placed further apart from the center of the calibration structure to a distance of $\rho=200$cm, while regarding the camera's height positions with respect to the ground floor, they were placed at heights 90cm (all at the same time) , 110cm (all at the same time), and 90cm and 110 cm interchangeably with respect to their spatial position. At each camera configuration, we performed 3 captures of the calibration structure, while rotating it around its $y$ axis for $\approx 20^\circ$ (6 cameras case) and $\approx 15^\circ$ (8 cameras case) in-between subsequent captures. This dataset contains 18 multi-view captures in total.

Finally, a third dataset (``cameras\_4'') was recorded, containing multi-view captures of the calibration structure captured by 4 cameras placed in a perimetrical setup around the structure. The cameras were placed at $\rho=200$cm while their height positions, with respect to the ground floor, were 90, 100 and 130 cm. A single capture of the structure was performed in each of the height configurations. This dataset contains 3 multi-view captures in total.

\textbf{Metrics.} 
Given the camera pose estimation parameters extracted by any of the evaluated methods and the captured depthmaps of the calibration structure, we align each viewpoint's point cloud to the GCS. Let $dist(\pazocal{A},\pazocal{B})$ denote a distance metric between point clouds $\pazocal{A}$ and $\pazocal{B}$.
We define:
\begin{equation}
    \label{eq:RMS}
    dist(\pazocal{A},\pazocal{B}) = \sqrt{\frac{1}{N}\sum_{\mathbf{v} \in \pazocal{A}} \min_{\mathbf{u} \in \pazocal{B} } ||\mathbf{v}-\mathbf{u}||^2}
\end{equation}
as the root mean squared (RMS) distance between point clouds $\pazocal{A}$ and $\pazocal{B}$. In the case of the dataset ``arc'', the structure is not fully visible from the set of cameras, since there are areas of the space that are not covered by the cameras. If $\pazocal{A}$ denotes the captured registered point cloud and $\pazocal{B}$ denotes the point cloud of structure's virtual 3D model, for this dataset we report $d_1 = dist(\pazocal{A},\pazocal{B})$ as defined above.

For the other datasets (``cameras\_6\_8'' and ``cameras\_4'') we use a Hausdorff-like RMS metric, ie: $d_2 = \max(dist(\pazocal{A},\pazocal{B}),dist(\pazocal{B},\pazocal{A}))$. For brevity, in subsequent table results (i.e Tables \ref{tbl:DatasetArc} and \ref{tbl:DatasetCameras468}) we will refer to the two different metrics $d_1$ and $d_2$ simply, by RMS, while the exact meaning can be inferred by the respective dataset and the details that we provided in this paragraph.

\textbf{Evaluation Methodology.} We take a two-step approach in evaluating the performance of the studied methods in the task of camera pose estimation. First, we evaluate the initial alignment of the viewpoints using the procedure described in the metrics subsection. Subsequently, similar to \cite{sterzentsenko2018low}, we use the graph-based dense optimization of \cite{wscg2018}, to obtain a global, refined, solution using ICP, formulated with a point-to-plane error. We use the same metric as described in the respective section to assess the alignment of the point clouds after the dense optimization.

\begin{figure}[!tp]
\begin{center}
   \includegraphics[width=0.95\columnwidth]{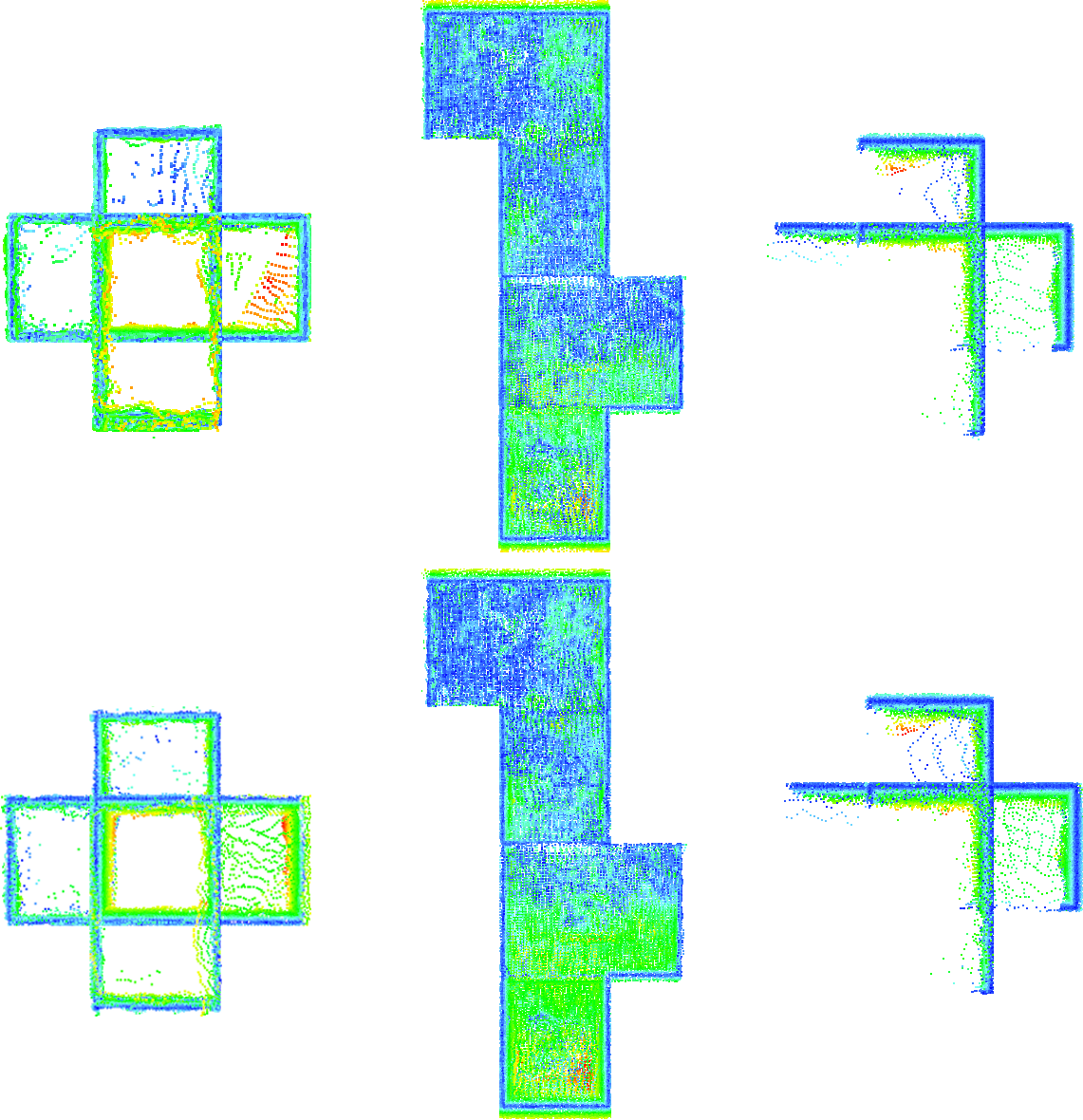}
\end{center}
   \vspace{-0.2in}
   \caption{Calibration results evaluated on \textit{arc} and \textit{cameras\_6\_8} datasets, marker-based (top) and our work (bottom). Depicted samples are (from left to right) 8 cameras in a full 360 setup (top and frontal views at 2 first columns) and \textit{arc} configuration.}
\label{fig:eight_and_arc}
\vspace{-0.2in}
\end{figure}

\textbf{Results.}
As depicted in Tables~\ref{tbl:DatasetArc} and~\ref{tbl:DatasetCameras468} our method performs an initial camera pose estimation which is inferior to the rest the SoA methods. However, despite its inferiority, the estimates it provides are good enough for the dense optimization algorithm to always achieve camera pose estimation performance competitive to the SoA marker-based method. Qualitative comparisons and visualization of quantitative errors are given in Fig.~\ref{fig:four_top} and Fig.~\ref{fig:eight_and_arc}. Each figure showcases, in color coding, a visualization of how each point of the aligned point cloud contributes to the overall distance between the aligned point cloud and the point cloud of the virtual model.

Furthermore, the proposed method achieved higher robustness compared to \cite{sterzentsenko2018low} as it did not fail to estimate the initial camera poses in any of datasets' samples. Contrariwise, as depicted in Fig.~\ref{fig:four_top}, the method of \cite{sterzentsenko2018low} had a total failure case to estimate camera poses. Additionally, the proposed method is markerless, as \cite{sterzentsenko2018low}, while not being bound to any restrictions in camera placement configurations or number of sensors in the multi-view camera setup. Thus, from the aforementioned discussion it becomes apparent that the proposed method fairly competes the SoA methods.

%% file: Conclusion.tex
Summarizing, in this work a differentiable variant of the Procrustes analysis was introduced, which efficiently bridged the intermediate semantic segmentation task with the targeted pose estimation one.
After an extensive evaluation the presented technique was assessed to enable more robust learned semantic correspondences to drive the initial pose estimates for volumetric sensor alignment.
Our work further increases the efficacy of markerless multi-sensor calibration, taking a step towards even larger flexibility while preserving user friendliness.
Nonetheless, our end-to-end formulation of this soft Procrustes analysis can be beneficial to the wider pose estimation task which is now moving towards dense key-point estimation, instead of purely regression the pose estimates.
We hope that our work stimulates further research in this direction as well as aid in easily calibrating multiple sensors.